\documentclass[conference]{IEEEtran2}

\usepackage{amsmath}
\usepackage{amsfonts}
\usepackage{amssymb}
\usepackage{amsbsy}
\usepackage{graphicx}
\usepackage{times}
\usepackage{default}
\usepackage{color}
\usepackage[noadjust]{cite}
\usepackage{enumerate}
\usepackage{tikz}
\usepackage{subcaption}
\usepackage{amsthm}

\newcommand{\Rd}{\mathbb{R}^d}

\newcommand{\RR}{\mathbb{R}}

\newcommand{\QQ}{\mathcal{Q}}

\newcommand{\la}{\lambda}

\newcommand{\VV}{\Vert}

\DeclareMathOperator{\dx}{\text{d}x}
\DeclareMathOperator{\du}{\text{d}u}

\DeclareMathOperator{\supp}{supp}

\def\ba#1\ea{\begin{align*}#1\end{align*}}	
\def\ban#1\ean{\begin{align}#1\end{align}}	
\def\bac#1\eac{\vspace{\abovedisplayskip}{\par\centering$\begin{aligned}#1\end{aligned}$\par}\addvspace{\belowdisplayskip}}	

\newtheorem{theorem}{Theorem}

\newtheorem{definition}{Definition}
\newtheorem{proposition}{Proposition}

\newtheorem{remark}{Remark}
\IEEEoverridecommandlockouts 
\newtheorem*{SL}{Schur's Lemma}
\title{Deep Convolutional Neural Networks Based on \\ Semi-Discrete Frames}
\author{
\IEEEauthorblockN{Thomas Wiatowski and  Helmut B\"olcskei}\vspace{.2cm}
\IEEEauthorblockA{
       Dept.~IT \& EE, ETH Zurich, Switzerland\\
         {Email: \{withomas,\ boelcskei\}@nari.ee.ethz.ch} 
}}

\begin{document}
\maketitle

\begin{abstract} Deep convolutional neural networks have led to breakthrough results in practical feature extraction applications. The mathematical analysis of these networks was pioneered by Mallat \cite{MallatS}. Specifically, Mallat considered so-called scattering networks based on identical semi-discrete wavelet frames in each network layer, and proved translation-invariance as well as deformation stability of the resulting feature extractor. The purpose of this paper is to develop Mallat's theory further by allowing for different and, most importantly, general semi-discrete frames (such as, e.g., Gabor frames, wavelets, curvelets, shearlets, ridgelets) in distinct network layers. This allows to extract wider classes of features than point singularities resolved by the wavelet transform. Our generalized feature extractor is proven to be translation-invariant, and we develop deformation stability results for a larger class of deformations than those considered by Mallat. For Mallat's wavelet-based feature extractor, we get rid of a number of technical conditions. The mathematical engine behind our results is continuous frame theory, which allows us to completely detach the invariance and deformation stability proofs from the particular algebraic structure of the underlying frames.
\end{abstract}

\section{Introduction}
A central task in signal classification is feature extraction \cite{Bengio}. For example, we may want to detect whether an \-image contains a certain handwritten digit \cite{MNIST}. Moreover, this should be possible independently of the feature's spatial (or temporal) location within the signal, which motivates the use of translation-invariant feature extractors. In addition, sticking to the example of handwritten digits, we want the feature extractor to be robust with respect to different handwriting styles. This is typically accounted for by asking for stability with respect to non-linear deformations of the feature to be extracted.

Spectacular success in many practical classification tasks has been reported for feature extractors generated by deep convolutional neural networks \cite{LeCun,Hinton}. The mathematical ana\-lysis of such networks was initiated by Mallat in \cite{MallatS}. Mallat's theory applies to so-called scattering networks, where signals are propagated through layers that compute the modulus of wavelet coefficients. The resulting feature extractor is pro\-vably translation-invariant and stable with respect to certain non-linear deformations. Moreover, it leads to state-of-the-art results in various image classification tasks \cite{Bruna,BrunaCSO}. 

The wavelet transform resolves signal features cha\-racterized by point singularities, but is not very effective in dealing with signals dominated by anisotropic features, such as, e.g., edges in images \cite{Shearlets}.  It thus seems natural to ask whether Mallat's theory on scattering networks can be extended to general signal transformations. Moreover, certain audio classification problems \cite{Anden} suggest that scattering networks with different signal transformations in different layers would be desirable in practice. 

\paragraph*{Contributions}
The goal of this paper is to extend Mallat's theory to cope with ge\-neral signal transformations (e.g., Gabor frames, wavelets, curvelets, shearlets, ridgelets), as well as to allow different signal transformations in different layers of the network, all that while retaining translation-invariance and deformation stability. Our second major contribution is a new deformation stability bound valid for a class of non-linear deformations that is wider than that considered by Mallat in \cite{MallatS}. The proofs in \cite{MallatS} all hinge critically on the wavelet transform's structural properties, whereas the technical arguments in our proofs are completely detached from the particular structure of the signal transforms. This leads to simplified and shorter proofs for translation-invariance and deformation stability. Moreover, in the case of Mallat's wavelet-based feature extractor we show that the admissibility condition for the mother wavelet (defined in \cite[Theorem 2.6]{MallatS}) is not needed. The mathe\-matical engine behind our results is the theory of continuous frames \cite{Antoine}. 

\paragraph*{Notation and preparatory material}
The complex conjugate of $z \in \mathbb{C}$ is denoted by $\overline{z}$. The Euclidean inner product of $x,y \in \mathbb{C}^d$ is $\langle x, y \rangle:=\sum_{i=1}^{d}x_i \overline{y_i}$, with associated norm $|x|:=\sqrt{\langle x, x \rangle}$. The supremum norm of a matrix $M\in \mathbb{R}^{d\times d}$ is defined by $|M|_{\infty}:=\sup_{i,j}|M_{i,j}|$, and the supremum norm of a tensor $T\in \mathbb{R}^{d\times d \times d}$ is $|T|_{\infty}:=\sup_{i,j,k}|T_{i,j,k}|$. We write $B_R(x)\subseteq \Rd$ for the open ball of radius $R>0$ centered at $x\in \Rd$. The Borel $\sigma$-algebra of $\Rd$ is denoted by $\mathbb{B}$. For a $\mathbb{B}$-measurable function $f:\Rd \to \mathbb{C}$, we write $\int_{\Rd} f(x) \mathrm dx$ for the integral of $f$ with respect to Lebesgue measure $\mu_L$. For $p \in [1,\infty)$,  $L^p(\Rd)$ denotes the space of all $\mathbb{B}$-measurable functions $f:\Rd \to \mathbb{C}$ such that $\VV f \VV_p:= (\int_{\Rd}|f(x)|^p \mathrm dx)^{1/p}<\infty.$ For $f,g \in L^2(\Rd)$ we set $\langle f,g \rangle :=\int_{\Rd}f(x)\overline{g(x)}\mathrm dx$. The operator norm of the linear bounded operator $A:L^p(\Rd) \to L^q(\Rd)$ is designated by $\VV A\VV_{p,q}$. $\text{Id}:L^p(\Rd) \to L^p(\Rd)$ stands for the identity operator on $L^p(\Rd)$. For a countably infinite set $\QQ$, $(L^2(\Rd))^\QQ$ denotes the space of sets $s:=\{ f_q \}_{q\in \QQ}$, $f_q \in L^2(\Rd)$, $\forall q \in \QQ$, such that $||| s |||:=(\sum_{q \in \QQ} \VV f_q\VV_2^2)^{1/2}<\infty$. We write $\mathbf{S}(\Rd)$ for the Schwartz space, i.e., the space of functions $f:\Rd \to \mathbb{C}$ whose derivatives along with the function itself are rapidly decaying \cite[Section 7.3]{Rudin}. We denote the Fourier transform of $f \in L^1(\Rd)$ by $\widehat{f}(\omega):=\int_{\Rd}f(x)e^{-2\pi i \langle x,  \omega \rangle }\mathrm dx$, and extend it in the usual way to $L^2(\Rd)$ \cite[Theorem 7.9]{Rudin}. The convolution of $f\in L^2(\Rd)$ and $g\in L^1(\Rd)$ is $(f\ast g)(y):=\int_{\Rd}f(x)g(y-x)\mathrm dx$. We write $T_tf(x):=f(x-t)$, $t \in \Rd$, for the translation operator, and $M_\omega f(x):=e^{2\pi i \langle x , \omega\rangle }f(x)$, $\omega \in \Rd$, for the modulation operator. Involution is defined by $(If)(x):=\overline{f(-x)}$. We denote the gradient of a function $f:\Rd \to \mathbb{C}$ as $\nabla f$. For a vector field $v:\Rd \to \Rd$, we write $Dv$ for its Jacobian matrix, and $D^2v$  for its Jacobian tensor, with associated norms $\VV v \VV_\infty:=\sup_{x\in\Rd} |v(x)|$, $\VV D v \VV_\infty:=\sup_{x\in \Rd}|(Dv)(x)|_{\infty}$, and $\VV D^2 v \VV_\infty:=\sup_{x\in \Rd}|(D^2v)(x)|_{\infty}$. For a scalar field $w:\Rd \to \mathbb{C}$, we define the norm $\VV w \VV_\infty:=\sup_{x\in \Rd} |w(x)|$.

\section{Mallat's wavelet-based feature extractor}\label{architecture}
\begin{figure}[t!]
\centering
\begin{tikzpicture}[scale=0.9,level distance=11mm]
   				\tikzstyle{level 1}=[sibling distance=20mm]   
  				 \tikzstyle{level 2}=[sibling distance=5mm]
  				  \tikzstyle{level 3}=[sibling distance=1mm]
				    \tikzstyle{level 4}=[sibling distance=5mm]
  						   \node {$f$}
     							child {node {$\big|f\ast \psi_{\lambda^{(j)}}\big|$}
      							child {node {$\big|\big|f\ast \psi_{\lambda^{(j)}}\big|\ast \psi_{\lambda^{(l)}}\big|$}
							child {node {$\big|\big|\big|f\ast \psi_{\lambda^{(j)}}\big|\ast \psi_{\lambda^{(l)}}\big|								\ast \psi_{\lambda^{(m)}}\big|$}
							child {[fill] circle (2pt)}
							child {[fill] circle (2pt)}
							child {[fill] circle (2pt)}}
							child[fill=none] {edge from parent[draw=none]}         												child[fill=none] {edge from parent[draw=none]}
							}
    							child[fill=none] {edge from parent[draw=none]}
							child[fill=none] {edge from parent[draw=none]}
							}
							child {node {$\big|f\ast \psi_{\lambda^{(p)}}\big|$}
							child {[fill] circle (2pt)
							child {[fill] circle (2pt)
							child {[fill] circle (2pt)}
							child {[fill] circle (2pt)}
							child {[fill] circle (2pt)}
							}
							}
							}
     							child {node {$\big|f\ast \psi_{\lambda^{(q)}}\big|$}
							child[fill=none] {edge from parent[draw=none]}
							child[fill=none] {edge from parent[draw=none]}
       							child {node {$\big|\big|f\ast \psi_{\lambda^{(q)}}\big|\ast \psi_{\lambda								^{(r)}}\big|$
							}
							child[fill=none] {edge from parent[draw=none]}
							child[fill=none] {edge from parent[draw=none]}
							child {node {$\big|\big|\big|f\ast \psi_{\lambda^{(q)}}\big|\ast \psi_{\lambda								^{(r)}}\big|\ast \psi_{\lambda^{(s)}}\big|$}
							child {[fill] circle (2pt)}
							child {[fill] circle (2pt)}
							child {[fill] circle (2pt)}}
							}
     							};
			\end{tikzpicture}
			\caption{Scattering network architecture based on wavelet filtering.} 
\label{fig2}
\end{figure}
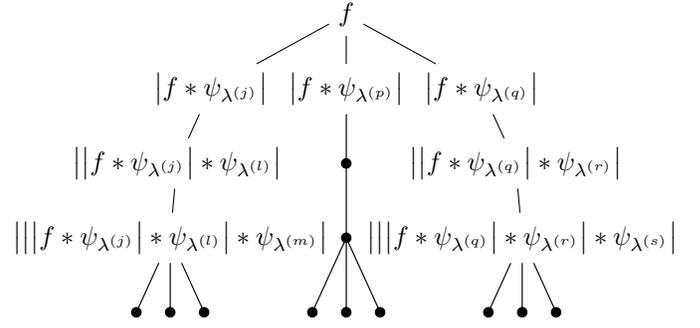

We set the stage by briefly reviewing Mallat's construction \cite{MallatS}. The basis for Mallat's feature extractor $\Phi_M$ is a multi-stage wavelet filtering technique followed by modulus operations. The extracted features $\Phi_M(f)$ of a signal $f\in L^2(\Rd)$ are defined as the set of low-pass filtered functions 
\begin{equation}\label{help11}
| \cdots | \ |f \ast \psi_{\lambda^{(l)}}| \ast  \psi_{\lambda^{(m)}}|\cdots  \ast \psi_{\lambda^{(n)}}|\ast \phi_J,
\end{equation}
labeled by the indices $\lambda^{(l)},\lambda^{(m)},\dots,\lambda^{(n)} \in \Lambda_W:=\big\{ (j,k) \ | \ j>-J, \ k \in \{ 1,\dots,K\} \big\}$ corresponding to pairs of scales and directions. The wavelets $\{ \psi_{\lambda} \}_{\la \in \Lambda_W}$ and the low-pass filter $\phi_J$ are atoms of a semi-discrete Parseval wavelet frame $\Psi_{\Lambda_W}$ and hence satisfy
\begin{equation*}\label{LPW}
\VV \phi_J \ast f \VV^2_2+  \sum_{\la \in \Lambda_W} \VV\psi_{\lambda} \ast f\VV_2^2=\VV f \VV_2^2, \hspace{0.3cm} \forall f\in L^2(\Rd).
\end{equation*}
We refer the reader to Appendix \ref{sec:sdf} for a short review of the theory of semi-discrete frames.
The architecture corresponding to \eqref{help11}, illustrated in Figure \ref{fig2}, is known as \textit{scattering network} \cite{Bruna}, and uses the same wavelets $\{ \psi_{\lambda} \}_{\lambda\in \Lambda_W}$ in every network layer.

It is shown in \cite{MallatS} that the feature extractor  $\Phi_M$ in \eqref{help11} is translation-invariant, in the sense that
$$
\Phi_M(T_t f)=T_t \Phi_M(f), \hspace{0.5cm} \forall t \in \Rd, \ \forall f\in L^2(\Rd),
$$
where $T_t$ is applied element-wise in $T_t \Phi_M(f)$. Further, it is proved in \cite{MallatS} that $\Phi_M$ is stable with respect to deformations of the form \begin{equation}\label{hoihoihoi}F_{\tau}f(x):=f(x-\tau(x)).\end{equation} Specifically, for the normed function space $(H_M,\VV \cdot \VV_{H_M})$ defined in \eqref{HMN} below, Mallat proved that there 
exists a constant $C>0$ such that for all $f\in H_M$ and every $\tau \in C^2(\Rd,\Rd)$ with\footnote{It is actually the assumption $\VV D \tau \VV_\infty\leq\frac{1}{2d}$, rather than $\VV D \tau \VV_\infty\leq\frac{1}{2}$ as stated in \cite[Theorem 2.12]{MallatS}, that is needed in \cite[Eq. E.31]{MallatS} to establish $|\det(\text{Id}- D \tau(x))| \geq 1- d\VV D \tau \VV_\infty\geq 1/2$. 
}
 $\VV D \tau \VV_\infty\leq\frac{1}{2d}$, the deformation error satisfies 
\begin{equation}\label{mallatbound}
\begin{split}
||| \Phi_M(f) &-\Phi_M(F_{\tau} f)|||\leq\\
&C\big(2^{-J}\VV \tau \VV_\infty + J\VV D \tau \VV_\infty + \VV D^2 \tau \VV_\infty\big)\VV f \VV_{H_M}.
\end{split}
\end{equation}

\section{Generalized feature extractor}\label{properties}
\begin{figure}[t!]
\centering
\begin{tikzpicture}[scale=0.9,level distance=11mm]
   				\tikzstyle{level 1}=[sibling distance=20mm]   
  				 \tikzstyle{level 2}=[sibling distance=5mm]
  				  \tikzstyle{level 3}=[sibling distance=1mm]
				    \tikzstyle{level 4}=[sibling distance=5mm]
  						   \node {$f$}
     							child {node {$\big|f\ast f_{\lambda_1^{(j)}}\big|$}
      							child {node {$\big|\big|f\ast f_{\lambda_1^{(j)}}\big|\ast f_{\lambda_{2}^{(l)}}\big|$}
							child {node {$\big|\big|\big|f\ast f_{\lambda_1^{(j)}}\big|\ast f_{\lambda_{2}^{(l)}}\big|\ast f_{\lambda_3^{(m)}}\big|$}
							child {[fill] circle (2pt)}
							child {[fill] circle (2pt)}
							child {[fill] circle (2pt)}}
							child[fill=none] {edge from parent[draw=none]}         												child[fill=none] {edge from parent[draw=none]}
							}
    							child[fill=none] {edge from parent[draw=none]}
							child[fill=none] {edge from parent[draw=none]}
							}
							child {node {$\big|f\ast f_{\lambda_1^{(p)}}\big|$}
							child {[fill] circle (2pt)
							child {[fill] circle (2pt)
							child {[fill] circle (2pt)}
							child {[fill] circle (2pt)}
							child {[fill] circle (2pt)}
							}
							}
							}
     							child {node {$\big|f\ast f_{\lambda_1^{(q)}}\big|$}
							child[fill=none] {edge from parent[draw=none]}
							child[fill=none] {edge from parent[draw=none]}
       							child {node {$\big|\big|f\ast f_{\lambda_1^{(q)}}\big|\ast f_{\lambda_{2}								^{(r)}}\big|$
							}
							child[fill=none] {edge from parent[draw=none]}
							child[fill=none] {edge from parent[draw=none]}
							child {node {$\big|\big|\big|f\ast f_{\lambda_1^{(q)}}\big|\ast f_{\lambda_{2}								^{(r)}}\big|\ast f_{\lambda_{3}^{(s)}}\big|$}
							child {[fill] circle (2pt)}
							child {[fill] circle (2pt)}
							child {[fill] circle (2pt)}}
							}
     							};
			\end{tikzpicture}
			\caption{Scattering network architecture based on general multi-stage filtering \eqref{a}. The function $f_{\lambda_{n}^{(k)}}$ is the $k$-th atom of the semi-discrete frame $\Psi_n$ associated with the $n$-th layer.} 
\label{fig1}
\end{figure}
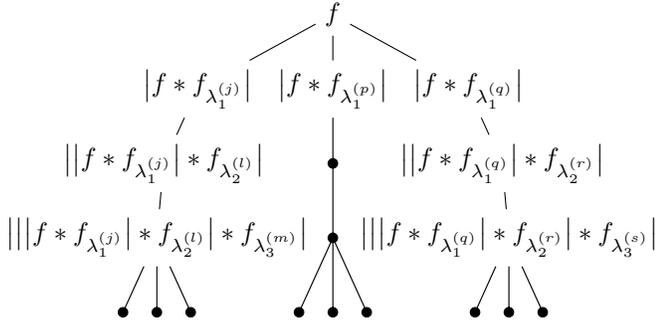

In this section, we describe our generalized feature extractor and start by introducing the notion of a \textit{frame collection}. 
\begin{definition} For all $n \in \mathbb{N}$, let $\Psi_{n}$ be a semi-discrete frame with frame bounds $A_n, B_n>0$ and atoms $\{ f_{\la'_n}\}_{\la'_n \in \Lambda'_n} \subseteq L^1(\Rd) \cap L^2(\Rd)$ indexed by a countable set $\Lambda'_n$. The sequence $\Psi:=( \Psi_n )_{n\in \mathbb{N}}$ is called a frame collection with frame bounds $A=\inf_{n \in \mathbb{N}}A_n$ and  $B=\sup_{n \in \mathbb{N}}B_n.$
\end{definition}
The elements $\Psi_n$, $n \in \mathbb{N}$, in a frame collection correspond to particular layers in the generalized scattering network defined below. In Mallat's construction one atom of the semi-discrete wavelet frame $\Psi_{\Lambda_W}$, namely the low-pass filter $\phi_J$, is singled out to generate the output set  \eqref{help11} of the feature extractor $\Phi_M$. We honor Mallat's terminology and designate one of the atoms  $\{ f_{\la'_n}\}_{\la'_n \in \Lambda'_n}$ of each frame $\Psi_n$ in the frame collection $\Psi$ as \textit{output-generating atom}. Note, however, that our theory does not require this atom to have low-pass characteristics. Specifically, we set $\phi_n:=f_{\lambda^\ast_n}$ for an arbitrary, but fixed $\lambda^\ast_n \in \Lambda'_n$. From now on, we therefore write 
$$\{ \phi_n \}\cup\{ f_{\la_n}\}_{\la_n \in \Lambda_n}, \hspace{0.5cm}\Lambda_n:=\Lambda'_n \backslash \{ \lambda_n^\ast\},$$ for the atoms of the semi-discrete frame $\Psi_n$. The reader might want to think of the discrete index set $\Lambda_n$  as a collection of scales, directions, or frequency-shifts.
\begin{remark}
Examples of structured frames that satisfy the ge\-neral semi-discrete frame condition \eqref{condii} and will hence be seen, in Theorem \ref{mainmain}, to be applicable in the construction of generalized feature extractors are, e.g., Gabor frames \cite{Groechening}, curvelets \cite{CandesDonoho2,Grohs_alpha}, shearlets \cite{Shearlets}, ridgelets \cite{Ridgelet,Grohs_transport}, and, of course, wavelets \cite{MallatW} as considered by Mallat in \cite{MallatS}.
\end{remark}
We now introduce our generalized scattering network. To this end, we generalize the multi-stage filtering technique underlying Mallat's scattering network to allow for general semi-discrete frames that can, in addition, be different in different layers. This requires the definition of a general modulus-convolution ope\-rator, and of paths on index sets.
\begin{definition} Let $\Psi=( \Psi_{n})_{n \in \mathbb{N}}$ be a frame collection with atoms $\{ \phi_n \}\cup\{ f_{\la_n}\}_{\la_n \in \Lambda_n}$. For $1\leq m< \infty$, define the set $\Lambda_1^m:=\Lambda_1\times \Lambda_2\times \dots \times \Lambda_m$. An ordered sequence $q=(\lambda_1,\lambda_{2},\dots, \lambda_m) \in \Lambda_1^m$ is called a path. The empty path, $e:=\emptyset$, defines the set $\Lambda_1^0:=\{ e \}$. 
The modulus-convolution operator is defined as $U:\big(\bigcup_{k=1}^\infty \Lambda_k \big)\times L^2(\Rd) \to L^2(\Rd)$, $ U(\lambda_n,f):=U[\lambda_n]f:=|f \ast f_{\lambda_n}|,$
where $f_{\la_n} \in L^1(\Rd) \cap L^2(\Rd)$ are the atoms of the semi-discrete frame $\Psi_n$ associated with the $n$-th layer in the network.
\end{definition}
We also need to extend the operator $U$ to paths $q \in \Lambda_1^m$ and do that according to
\begin{equation}\label{a}
\begin{split}
U[q]f:&=U[\lambda_m] \cdots U[\lambda_{2}]U[\lambda_{1}]f\\
&=|\cdots ||f \ast f_{\lambda_{1}}| \ast  f_{\lambda_{2}}|\cdots  \ast f_{\lambda_m}|,
\end{split}
\end{equation}
where we set $U[e]f=f$. Note that the multi-stage filtering opera\-tion \eqref{a} is well-defined, as $\VV U[q] f \VV_2 \leq \big(\prod_{n=1}^{m}\VV f_{\la_n}\VV_1\big) \VV f \VV_2$,  thanks to Young's inequality \cite[Theorem 1.2.12]{Grafakos}. Figure \ref{fig1} illustrates the gene\-ralized scattering network with different semi-discrete frames in different layers. 

We can now put the pieces together and define the gene\-ralized feature extractor $\Phi_\Psi$.
\begin{definition}\label{defn2}
Let $\Psi=( \Psi_{n})_{n \in \mathbb{N}}$ be a frame collection, and define $\QQ:=  \bigcup_{k=0}^\infty \Lambda_{1}^k$. Given a path $q \in \Lambda_1^n$, $n \geq 0$,  we write $\phi[q]:=\phi_{n+1}$ for the output-generating atom of the semi-discrete frame $\Psi_{n+1}$. The feature extractor $\Phi_\Psi$ with respect to the frame collection $\Psi$ is defined as
\begin{equation}\label{ST}
\Phi_\Psi (f):=\{ U[q]f \ast \phi[q] \}_{q \in \QQ}.
\end{equation}
\end{definition}
\section{Main result}\label{main}
The main result of this paper is the following theorem, sta\-ting that the feature extractor $\Phi_\Psi$ defined in \eqref{ST} is translation-invariant and stable with respect to time-frequency deformations of the form
\begin{equation}\label{def:deform}
F_{\tau,\omega} f(x):=e^{2\pi i \omega(x)}f(x-\tau(x)).
\end{equation}
The class of deformations we consider is wider than the one in Mallat's theory, who considered translation-like deformations of the form $f(x-\tau(x))$ only. Modulation-like deformations $e^{2\pi i \omega(x)}f(x)$ occur, e.g., if we have access only to a band-pass version of the signal $f \in L^2(\Rd)$.
\begin{theorem}\label{mainmain}
 Let $\Psi$ be a frame collection with upper frame bound $B\leq 1$. The feature extractor $\Phi_\Psi$ defined in \eqref{ST} is translation-invariant. Further, for $R>0$, define the space of $R$-band-limited functions
$$
H_R:=\{ f \in L^2(\Rd) \ | \ \supp(\hat{f}) \subseteq B_{R}(0)\}.
$$Then, the feature extractor $\Phi_\Psi$ is stable on $H_R$ with respect to non-linear deformations \eqref{def:deform}, i.e., there exists $C>0$ (that does not depend on $\Psi$) such that for all $f \in H_R$ and all $\omega \in C(\Rd,\RR)$, $\tau \in C^1(\Rd,\Rd)$ with $\VV D \tau \VV_\infty\leq\frac{1}{2d}$, it holds that
\begin{equation}\label{mainmainmain}
||| \Phi_\Psi(f)-\Phi_\Psi(F_{\tau,\omega} f) |||\leq C \big( R\VV \tau \VV_\infty + \VV \omega\VV_\infty\big) \VV f \VV_2.
\end{equation}
\end{theorem}
The proof of Theorem $\ref{mainmain}$ can be found in Appendix \ref{Appendix}. Our main result shows that translation-invariance and deformation stability are retained for the generalized feature extractor $\Phi_\Psi$. The strength of this result derives itself from the fact that the only condition on $\Psi$ for this to hold is $B\leq1$. This condition is easily met by normalizing the frame elements accordingly. Such a normalization impacts neither translation-invariance nor the constant $C$ in \eqref{mainmainmain} which is seen, in \eqref{stab_const}, to be independent of $\Psi$. All this is thanks to our proof techniques, unlike those in \cite{MallatS}, being independent of the algebraic structure of the underlying frames. This is accomplished through a genera\-lization of a Lipschitz-continuity result by Mallat \cite[Proposition 2.5]{MallatS} for the feature extractor $\Phi_\Psi$ (stated in Proposition \ref{summary} in Appendix \ref{Appendix}), and by employing a partition of unity argument \cite{RudinW} for band-limited functions. 
\section{Relation to Mallat's results}\label{discussion}
To see how Mallat's wavelet-based architecture is covered by our Theorem \ref{mainmain}, simply note that by \cite[Eq. 2.7]{MallatS} the atoms $\{ \phi_J \}\cup \{ \psi_{\la} \}_{\la \in \Lambda_W}$ satisfy \eqref{freqcover} with $A=B=1$. Since Mallat's construction uses the same wavelet frame in each layer, this trivially implies $\sup_{n \in \mathbb{N}}B_n \leq1$. 

Mallat imposes additional technical conditions on the atoms $\{ \phi_J \}\cup \{ \psi_{\la} \}_{\la \in \Lambda_W}$, one of which is the so-called scattering admissibility condition for the mother wavelet, defined in \cite[Theorem 2.6]{MallatS}. To the best of our knowledge, no wavelet in $\Rd$, $d\geq 2$, satisfying this condition has been reported in the literature. 

Mallat's stability bound \eqref{mallatbound} applies to signals $f\in L^2(\Rd)$ satisfying 
\begin{equation}\label{HMN}
\VV f \VV_{H_M}:=\sum_{m=0}^\infty\sum_{q \in {\Lambda_W}_1^m} \VV U[q]f\VV_2<\infty.
\end{equation} 
While \cite[Section 2.5]{MallatS} cites numerical evidence on \eqref{HMN} being finite for a large class of functions $f \in L^2(\Rd)$, it seems difficult to establish this analytically.

Finally, the stability bound \eqref{mallatbound} depends on the parameter $J$, which determines the coarsest scale resolved by the wavelets $\{ \psi_\la \}_{\la \in \Lambda_W}$. For $J \to \infty$ the term $2^{-J}\VV \tau \VV_\infty$ vanishes; however, the term $J\VV D \tau \VV_\infty$ tends to infinity.

Our main result shows that i) the scattering admissibility condition in \cite{MallatS} is not needed, ii) instead of the signal class characterized by \eqref{HMN} our result applies \textit{provably} to the space of $R$-band-limited functions $H_R$, and iii) our deformation stability bound \eqref{mainmainmain}, when particularized to wavelets, besides applying to a wider class of non-linear deformations, namely \eqref{def:deform} instead of \eqref{hoihoihoi}, is independent of $J$.

The proof technique used in \cite{MallatS} to establish \eqref{mallatbound} makes heavy use of structural specifics of the atoms $\{ \phi_J \}\cup \{ \psi_{\la} \}_{\la \in \Lambda_W}$, namely isotropic dilations, va\-nishing moment conditions, and a constant number $K\in \mathbb{N}$ of directional wavelets across scales. 

\appendices
\section{Semi-discrete frames}\label{sec:sdf}
This appendix gives a short review of semi-discrete frames \cite{Antoine}.
\begin{definition}\label{defn:contframe}
Let  $\{ f_\la \}_{\la \in \Lambda}\subseteq L^1(\Rd)\cap L^2(\Rd)$  be a set of functions indexed by a countable set $\Lambda$. The set of translated and involuted functions $$\Psi_\Lambda=\{ T_b If_\la\}_{(\la,b) \in \Lambda \times \Rd}$$ is called a semi-discrete frame, if there exist constants $A,B >0$ such that 
\begin{equation}\label{condii}
A \VV f \VV_2^2 \leq \sum_{\lambda\in \Lambda} \VV f \ast f_{\lambda} \VV_2^2 \leq B \VV f \VV_2^2
\end{equation}
for all $f\in L^2(\Rd)$. The functions $\{ f_\la \}_{\la \in \Lambda}$ are called the atoms of the semi-discrete frame $\Psi_\Lambda$. When $A=B$ the semi-discrete frame is said to be tight. A tight semi-discrete frame with frame bound $A=1$ is called a semi-discrete Parseval frame.
\end{definition}
The frame operator associated with the semi-discrete frame $\Psi_\Lambda$ is defined in the weak sense by $S_{\Lambda}:L^2(\Rd) \to L^2(\Rd)$,
\begin{equation*}\label{eq:sdfo}
S_{\Lambda}f= \Big( \sum_{\la \in \Lambda} f_\la \ast I f_\la\Big)\ast f.
\end{equation*}
$S_\Lambda$ is a bounded, positive, and boundedly invertible operator \cite{Antoine}.

The reader might want to think of semi-discrete frames as shift-invariant frames \cite{Jansen}, where the translation parameter is left unsampled. The discrete index set $\Lambda$ typically labels a collection of scales, directions, or frequency-shifts. 
For instance, as illustrated in Section \ref{architecture}, Mallat's scattering network is based on a semi-discrete Parseval frame of directional wavelet structure, where the atoms $\{\phi_J \} \cup \{ \psi_\la\}_{ \la\in \Lambda_W}$ are indexed by the set $\Lambda_W=\big\{ (j,k) | \ j>-J, \ k\in \{1,\dots,K \} \big\}$, labeling a collection of scales and directions.

For shift-invariant frames it is often convenient to work with a unitarily equivalent representation of the frame operator.
\begin{proposition}\cite[Theorem 5.11]{MallatW}\label{freqcoverthm2}
Let $\Lambda$ be a countable index set. The functions $\{ f_\la \}_{\la \in \Lambda}\subseteq L^1(\Rd) \cap L^2(\Rd)$ are atoms of the semi-discrete frame $\Psi_\Lambda= \{ T_b If_\la\}_{(\la,b) \in \Lambda\times \Rd}$ with frame bounds $A,B>0$ if and only if
\begin{equation}\label{freqcover}
A\leq \sum_{\lambda \in \Lambda} |\widehat{f_{\lambda}}(\omega)|^2\leq B, \hspace{0.5cm} a.e. \ \omega \in \Rd.
\end{equation}
\end{proposition}
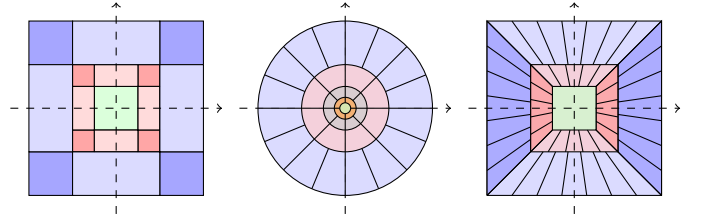
\begin{figure}[t]
\begin{center}
\begin{tikzpicture}
	\begin{scope}[scale=.145]
		\pgfsetfillopacity{.7}
	\filldraw[fill=green!20!white, draw=black]
	(-19,-2)--(-19,2) -- (-23,2) -- (-23,-2) -- (-19,-2);
	
	\filldraw[fill=red!50!white, draw=black]
	(-23,-2) -- (-23,-4) -- (-25,-4) -- (-25,-2) --(-23,-2);
	\filldraw[fill=red!50!white, draw=black]
	(-23,2) -- (-23,4) -- (-25,4) -- (-25,2) --(-23,2);
	\filldraw[fill=red!50!white, draw=black]
	(-19,-2) -- (-19,-4)--(-17,-4) -- (-17,-2) -- (-19,-2) ;
	\filldraw[fill=red!50!white, draw=black]
	(-19,2) -- (-19,4)--(-17,4) -- (-17,2) -- (-19,2) ;
	
 	\filldraw[fill=blue!50!white, draw=black]	
	(-17,-4)-- (-17,-8)--(-13,-8)--(-13,-4)--(-17,-4);
 	\filldraw[fill=blue!50!white, draw=black]	
	(-17,4)-- (-17,8)--(-13,8)--(-13,4)--(-17,4);
 	\filldraw[fill=blue!50!white, draw=black]	
	(-25,-4)--(-25,-8) -- (-29,-8) -- (-29,-4) -- (-25,-4);
 	\filldraw[fill=blue!50!white, draw=black]	
	(-25,4)--(-25,8) -- (-29,8) -- (-29,4) -- (-25,4);

 	\filldraw[fill=red!20!white, draw=black]	
	(-19,-2)--(-19,-4) -- (-23,-4)--(-23,-2)--(-19,-2);
 	\filldraw[fill=red!20!white, draw=black]	
	(-19,2)--(-19,4) -- (-23,4)--(-23,2)--(-19,2);
 	\filldraw[fill=red!20!white, draw=black]	
	(-17,-2) -- (-17,2)-- (-19,2)--(-19,-2)--(-17,-2);
 	\filldraw[fill=red!20!white, draw=black]	
	(-23,-2)--(-23,2)--(-25,2) --(-25,-2) --(-23,-2);

	\filldraw[fill=blue!20!white, draw=black]
	(-17,-4) -- (-17,-8) -- (-25,-8) -- (-25,-4) --(-17,-4);
	\filldraw[fill=blue!20!white, draw=black]
	(-17,4) -- (-17,8) -- (-25,8) -- (-25,4) --(-17,4);
	\filldraw[fill=blue!20!white, draw=black]
	(-17,-4) -- (-17,4) -- (-13,4) -- (-13,-4) -- (-17,-4);
	\filldraw[fill=blue!20!white, draw=black]
	(-25,-4) -- (-25,4) -- (-29,4)-- (-29,-4) -- (-25,-4);	

	\draw[->,dashed] (-30.7,0) -- (-11.3,0) ;
	\draw[->,dashed] (-21,-9.7) -- (-21,9.7);

	\draw[->,dashed] (-9.7,0) -- (9.7,0) ;
	\draw[->,dashed] (0,-9.7) -- (0,9.7);	
	\filldraw[fill=blue!20!white, draw=black]
   		  (0,0) circle (8cm);
	\filldraw[fill=red!20!white, draw=black]
   		  (0,0) circle (4cm);
	\filldraw[fill=black!20!white, draw=black]
   		  (0,0) circle (2cm);
	\filldraw[fill=orange!70!white, draw=black]
   		  (0,0) circle (1cm);
	\filldraw[fill=green!20!white, draw=black]
   		  (0,0) circle (.5cm);

	\draw (0.5,0) -- (8,0);
	\draw (0,0.5) -- (0,8);
	\draw (-.5,0) -- (-8,0);
	\draw (0,-.5) -- (0, -8);
	
	\draw (0.7071,0.7071) -- (5.6569,5.6569);
	\draw (-0.7071,0.7071) -- (-5.6569,5.6569);
	\draw (-0.7071,-0.7071) -- (-5.6569,-5.6569);
	\draw (0.7071,-0.7071) -- (5.6569,-5.6569);
	
	\draw (3.6955,1.5307) -- (7.3910,3.0615);
	\draw (-3.6955,-1.5307) -- (-7.3910,-3.0615);
	\draw (3.6955,-1.5307) -- (7.3910,-3.0615);
	\draw (-3.6955,1.5307) -- (-7.3910,3.0615);
	\draw (1.5307,3.6955) -- (3.0615,7.3910);
	\draw (1.5307,-3.6955) -- (3.0615,-7.3910);
	\draw (-1.5307,3.6955) -- (-3.0615,7.3910);
	\draw (-1.5307,-3.6955) -- (-3.0615,-7.3910);

	\filldraw[fill=blue!20!white, draw=black]
   		  (13,-8) -- (13,8)--(29,8) -- (29,-8) -- (13,-8);	
	\filldraw[fill=blue!40!white, draw=black]
   		  (13,-8) -- (17,-4)--(17,4) -- (13,8) -- (13,-8);	 
	\filldraw[fill=blue!40!white, draw=black]
   		  (29,-8) -- (25,-4)--(25,4) -- (29,8) -- (29,-8);		   
	\filldraw[fill=red!20!white, draw=black]
   		  (17,-4) -- (17,4)--(25,4) -- (25,-4) -- (17,-4);
	\filldraw[fill=red!40!white, draw=black]
   		  (17,-4) -- (19,-2)--(19,2) -- (17,4) -- (17,-4);		
	\filldraw[fill=red!40!white, draw=black]
   		  (25,-4) -- (23,-2)--(23,2) -- (25,4) -- (25,-4);			    
	\filldraw[fill=green!20!white, draw=black]	
   		  (19,-2) -- (19,2)--(23,2) -- (23,-2) -- (19,-2);

	\draw (23,2) -- (29,8);
	\draw (23,-2) -- (29,-8);
	\draw (19,2) -- (13,8);
	\draw (19,-2) -- (13,-8);
	
	\draw (23,-1.2) -- (25,-2.4);			
	\draw (23,-0.4) -- (25,-.8);
	\draw (23,0.4) -- (25,.8);
	\draw (23,1.2) -- (25,2.4);
	
	\draw (19,-1.2) -- (17,-2.4);			
	\draw (19,-0.4) -- (17,-.8);
	\draw (19,0.4) -- (17,.8);
	\draw (19,1.2) -- (17,2.4);

	\draw (19.8,2) -- (18.6,4);
	\draw (20.6,2) -- (20.2,4);
	\draw (21.4,2) -- (21.8,4);
	\draw (22.2,2) -- (23.4,4);
	
	\draw (19.8,-2) -- (18.6,-4);
	\draw (20.6,-2) -- (20.2,-4);
	\draw (21.4,-2) -- (21.8,-4);
	\draw (22.2,-2) -- (23.4,-4);
	
	\draw (18.1429,4) -- (15.2857,8);
	\draw (19.2857,4) -- (17.5714,8);
	\draw (20.4286,4) -- (19.8571,8);
	\draw (21.5714,4) -- (22.1429,8);
	\draw (22.7143,4) -- (24.4286,8);
	\draw (23.8571,4) -- (26.7143,8);	

	\draw (18.1429,-4) -- (15.2857,-8);
	\draw (19.2857,-4) -- (17.5714,-8);
	\draw (20.4286,-4) -- (19.8571,-8);
	\draw (21.5714,-4) -- (22.1429,-8);
	\draw (22.7143,-4) -- (24.4286,-8);
	\draw (23.8571,-4) -- (26.7143,-8);	
	
	\draw (25,2.8571) -- (29,5.7143);
	\draw (25,1.7143) -- (29,3.4286);
	\draw (25,0.5714) -- (29,1.1429);
	\draw (25,-0.5714) -- (29,-1.1429);
	\draw (25,-1.7143) -- (29,-3.4286);
	\draw (25,-2.8571) -- (29,-5.7143);	
	
	\draw (17,2.8571) -- (13,5.7143);
	\draw (17,1.7143) -- (13,3.4286);
	\draw (17,0.5714) -- (13,1.1429);
	\draw (17,-0.5714) -- (13,-1.1429);
	\draw (17,-1.7143) -- (13,-3.4286);
	\draw (17,-2.8571) -- (13,-5.7143);

\draw[->,dashed] (11.3,0) -- (30.7,0) ;
	\draw[->,dashed] (21,-9.7) -- (21,9.7);		
		
\end{scope}
\end{tikzpicture}
\end{center}
\caption{Frequency plane partitions in $\RR^2$ induced by atoms $\{ \widehat{f_\la}\}_{\la \in \Lambda}$ of semi-discrete tensor wavelets (left), semi-discrete curvelets (center), and semi-discrete cone-adapted shearlets (right).}
\end{figure}
\section{Proof of Theorem \ref{mainmain}}\label{Appendix}
We first prove translation-invariance. Fix $f\in L^2(\Rd)$ and define $C[q]f:= U[q]f\ast \phi[q]$, $ \forall q \in \QQ$. By \eqref{ST} it follows that $\Phi_\Psi$ is translation-invariant if and only if 
\begin{equation}\label{help1111}
C[q](T_tf)=T_t(C[q]f),\hspace{0.5cm} \ \forall t \in \Rd, \ \forall q \in \QQ.
\end{equation}
Due to $C[q](T_tf)=U[q] (T_tf) \ast \phi[q]$ and $$T_t(C[q]f)=T_t\big(U[q]f\ast \phi[q]\big)=\big(T_t(U[q] f)\big) \ast \phi[q],$$
\eqref{help1111} holds if $U[q] (T_tf)=T_t(U[q] f)$, $\forall t \in \Rd$, $\forall q \in \QQ$. The proof is concluded by noting that $U[q]$ is translation-invariant thanks to \eqref{a} and $$U[\la_n](T_t f)= |(T_t f) \ast f_{\la_n}| = |T_t(f \ast f_{\la_n})|=T_t (U[\la_n]f),$$ for all $t \in \RR^d$, $\la_n \in \bigcup_{k=1}^\infty \Lambda_k$.

Let us now turn to the proof of deformation stability, which is based on two key ingredients, the first being a generalization of a Lipschitz-continuity result by Mallat \cite[Proposition 2.5]{MallatS}:
\begin{proposition}\label{summary} Let $\Psi$ be a frame collection with upper frame bound $B\leq 1$. The feature extractor $\Phi_\Psi:L^2(\Rd) \to (L^2(\Rd))^\QQ$  is a bounded, Lipschitz-continuous operator with Lipschitz constant $L=\sqrt{B}$, i.e.,  
\begin{equation*}\label{zzz}
||| \Phi_\Psi(f) -\Phi_\Psi(h)||| \leq \sqrt{B} \VV f-h \VV_2 \end{equation*} for all $f,h \in L^2(\Rd).$
\end{proposition}
The proof of Proposition \ref{summary} is not given here, as it essentially follows that of \cite[Proposition 2.5]{MallatS} with minor changes. 
We now apply Proposition \ref{summary} with $h:=F_{\tau,\omega}f$ and get
\begin{equation*}\label{ahoiahoi}
||| \Phi_\Psi(f) -\Phi_\Psi(F_{\tau,\omega} f)||| \leq \VV f-F_{\tau,\omega} f \VV_2
\end{equation*}
 for all $f \in L^2(\Rd)$. Here, we used $\sqrt{B}\leq1$, due to $B\leq1$, as well as $h=F_{\tau,\omega}f \in L^2(\Rd)$, which is thanks to
\begin{equation*}
\begin{split}
\VV h \VV_2^2&=\VV F_{\tau,\omega}(f)\VV_2^2 =\int_{\Rd}|f(x-\tau(x))|^2\mathrm dx
\leq 2\VV f\VV^2_2,
\end{split}
\end{equation*}
obtained through the change of variables $u=x-\tau(x)$, together with 
\begin{equation}\label{cokabs}
\frac{\mathrm du}{\mathrm dx}=|\det(\text{Id}-D \tau(x))| \geq 1-d \VV D \tau \VV_\infty\geq 1/2.
\end{equation}
The inequalities in \eqref{cokabs} hold thanks to \cite[Corollary 1]{Brent} and $\VV D\tau \VV_\infty \leq\frac{1}{2d}$, respectively. 
The second key ingredient of our proof is a partition of unity argument \cite{RudinW} for band-li\-mited functions used to derive an upper bound on $ \VV f-F_{\tau,\omega} f \VV_2$. We first determine a function $\gamma$ such that $f=f\ast \gamma$ for all $f \in H_R$. Consider $ \eta \in \mathbf{S}(\Rd)$ such that $\widehat{\eta}(\omega)=1$, $\forall \omega\in B_{1}(0)$. Setting $\gamma(x):=R^d\eta(Rx)$ yields $\widehat{\gamma}(\omega)=\widehat{\eta}(\omega/R)$. Thus, $\widehat{\gamma}(\omega)=1$, $\forall \omega \in B_{R}(0)$, as well as $\widehat{f}=\widehat{f} \widehat{\gamma}$ and $f=f\ast \gamma$ for all $f \in H_R$. Then, we define the operator 
$A_{\gamma}:L^2(\Rd) \to L^2(\Rd)$, $A_{\gamma}(f)=f\ast \gamma$. Note that $A_\gamma$ is well-defined as $\gamma \in \mathbf{S}(\Rd) \subseteq L^1(\Rd)$. We now get 
\begin{equation*}\label{ahoiahoihoi}
\begin{split}
\VV f-F_{\tau,\omega}f \VV_{2}&=\VV A_\gamma f-F_{\tau,\omega}A_{\gamma}f \VV_{2}\\
&\leq \VV A_\gamma - F_{\tau,\omega}A_{\gamma}\VV_{2,2} \VV f\VV_2
\end{split}
\end{equation*} for all $f\in H_R$. In order to bound the norm $\VV A_\gamma - F_{\tau,\omega}A_{\gamma}\VV_{2,2}$, we apply Schur's Lemma to the integral operator 
$F_{\tau,\omega} A_\gamma -A_\gamma$.
\begin{SL}\cite[App. I.1]{Grafakos}
Let $k:\Rd \times \Rd \to \mathbb{C}$ be a locally integrable function satisfying $ \sup_{u\in \Rd} \int_{\Rd}|k(x,u)|\mathrm dx \leq C$
and 
$\sup_{x \in \Rd}\int_{\Rd}|k(x,u)|\mathrm du \leq C.$ Then, the integral ope\-rator $K$ given by  
$K(f)(x)=\int_{\Rd} f(u)k(x,u)\mathrm du,$
is a bounded operator from $L^2(\Rd)$ to $L^2(\Rd)$ with norm $\VV K \VV_{2,2}\leq C$.
\end{SL}
From the identity
\begin{equation*}
\begin{split}
F_{\tau,\omega} A_\gamma(f)(x)&=e^{2\pi i \omega(x)}\int_{\Rd}  \gamma(x-\tau(x)-u)f(u)\mathrm du,\\
\end{split}
\end{equation*}
it follows that  $F_{\tau,\omega} A_{\gamma}-A_{\gamma}$ has the kernel function
$k(x,u):= e^{2\pi i \omega(x)}\gamma(x-\tau(x)-u)-\gamma(x-u)$, which is locally integrable thanks to $\gamma \in \mathbf{S}(\Rd)$ and $\tau \in \mathbb{C}(\Rd,\Rd)$. We next use a first-order Taylor expansion in order to bound $|k(x,u)|$. To this end, let $x,u\in \Rd$, and define $h^{x,u}:\RR \to \mathbb{C}$, as
$h^{x,u}(t)=e^{2\pi i t \omega(x)}\gamma(x-t\tau(x)-u)-\gamma(x-u)$.
It follows that $h^{x,u}(0)=0$ and $h^{x,u}(1)=k(x,u)$. Therefore, we have $h^{x,u}(t)=h^{x,u}(0)+\int_{0}^t (\frac{\mathrm d}{\mathrm dt}h^{x,u})(\lambda)\mathrm d\lambda$, $\forall t \in \RR$. The special choice $t=1$ yields $|k(x,u)|=|h^{x,u}(1)|\leq \int_{0}^1|(\frac{\mathrm d}{\mathrm dt}h^{x,u})(\lambda)|\mathrm d\lambda$ with 
\begin{equation*}
\begin{split}
\Big|\Big(\frac{\mathrm d}{\mathrm dt}h^{x,u}\Big)(\lambda)\Big|&\leq \big| \big\langle \nabla \gamma(x-\lambda \tau(x)-u), \tau(x)\big\rangle \big|\\
&+ | 2\pi \omega(x)\gamma(x-\lambda\tau(x)-u)|\\
& \leq\VV \tau \VV_\infty| \nabla \gamma(x-\lambda \tau(x)-u)|\\
&+ 2\pi \VV \omega\VV_\infty |\gamma(x-\lambda\tau(x)-u)|.
\end{split}
\end{equation*}
Thanks to $\gamma,\nabla \gamma \in \mathbf{S}(\Rd)$, and $\mu_L([0,1])=1<\infty$, we can apply Fubini's Theorem to get
\begin{equation*}
\begin{split}
\int_{\Rd} |k(x,u)|\mathrm du &\leq\VV \tau \VV_{\infty} \int_{0}^1\int_{\Rd} |\nabla \gamma(x-\lambda\tau(x)-u)|\mathrm \du \mathrm d\lambda\\
&+2\pi \VV \omega \VV_{\infty} \int_{0}^1\int_{\Rd} | \gamma(x-\lambda\tau(x)-u)| \mathrm \du \mathrm d\lambda\\
&\leq\VV \tau \VV_\infty \VV \nabla \gamma \VV_1 + 2\pi \VV \omega \VV_{\infty}\VV \gamma \VV_1\\
&=R\VV \tau \VV_\infty  \VV \nabla \eta \VV_1+ 2\pi \VV \omega \VV_{\infty}\VV \eta \VV_1.
\end{split}
\end{equation*}
Similarly, we obtain
\begin{equation*}
\begin{split}
\int_{\Rd} |k(x,u)|\mathrm dx &\leq\VV \tau \VV_{\infty} \int_{0}^1\int_{\Rd} |\nabla \gamma(x-\lambda\tau(x)-u)|\mathrm \dx \mathrm d\lambda\\
&+2\pi \VV \omega \VV_{\infty} \int_{0}^1\int_{\Rd} | \gamma(x-\lambda\tau(x)-u)|\mathrm \dx \mathrm d\lambda\\
&\leq 2\VV \tau \VV_\infty\VV \nabla \gamma \VV_1+ 4\pi \VV \omega \VV_\infty\VV \gamma \VV_1\\
&=2R\VV \tau \VV_\infty\VV \nabla \eta \VV_1+ 4\pi \VV \omega \VV_\infty\VV \eta \VV_1\\
\end{split}
\end{equation*}
by the change of variables $y=x-\lambda \tau(x) -u$, together with 
\begin{equation}\label{helpabc}
\frac{\mathrm dy}{\mathrm dx}=|\det(\text{Id}-\lambda D \tau(x))| \geq 1- \lambda d\VV D \tau \VV_\infty\geq 1/2.
\end{equation}
The inequalities in \eqref{helpabc} hold thanks to \cite[Corollary 1]{Brent}, $\VV D\tau \VV_\infty \leq\frac{1}{2d}$, and $\lambda \in [0,1]$. The proof is completed by setting
\begin{equation}\label{stab_const}
C:=\max\big\{ 2\VV \nabla \eta \VV_1, 4\pi \VV \eta \VV_1\big\}\big(R\VV \tau \VV_\infty +\VV \omega \VV_\infty\big).
\end{equation} 

\bibliographystyle{IEEEtran}
\bibliography{scatbib}

\end{document}